\documentclass{INTERSPEECH2023}
\interspeechcameraready 

\usepackage{amsmath}
\usepackage{amsfonts}
\usepackage{graphicx}
\usepackage{booktabs}
\usepackage{microtype}
\usepackage{amssymb}
\usepackage{subfig}
\usepackage{multirow}

\title{Knowledge Distillation from Non-streaming to Streaming ASR Encoder\\using Auxiliary Non-streaming Layer}
\name{Kyuhong Shim, Jinkyu Lee, Simyung Chang and Kyuwoong Hwang}
\address{Qualcomm AI Research${}^{\dagger}$, Qualcomm Korea YH, Seoul, Republic of Korea 
\thanks{${}^{\dagger}$Qualcomm AI Research is an initiative of Qualcomm Technologies, Inc.}}
\email{\{kshim, jinkyu, simychan, kyuwoong\}@qti.qualcomm.com}

\begin{document}
\maketitle
 
\begin{abstract}
Streaming automatic speech recognition (ASR) models are restricted from accessing future context, which results in worse performance compared to the non-streaming models.
To improve the performance of streaming ASR, knowledge distillation (KD) from the non-streaming to streaming model has been studied, mainly focusing on aligning the output token probabilities.
In this paper, we propose a layer-to-layer KD from the teacher encoder to the student encoder.
To ensure that features are extracted using the same context, we insert auxiliary non-streaming branches to the student and perform KD from the non-streaming teacher layer to the non-streaming auxiliary layer.
We design a special KD loss that leverages the autoregressive predictive coding (APC) mechanism to encourage the streaming model to predict unseen future contexts.
Experimental results show that the proposed method can significantly reduce the word error rate compared to previous token probability distillation methods.
\end{abstract}
\noindent\textbf{Index Terms}: streaming speech recognition, non-streaming speech recognition, knowledge distillation



\section{Introduction}\label{sec:intro}

Streaming automatic speech recognition (ASR) is a fundamental technology used in various speech-related applications, such as live captioning, voice command recognition, and voice chat.
However, streaming ASR models face significant limitations because of their typical target systems' requirements for on-device and real-time speech recognition.
The primary challenge is that the model cannot access the future context in order to minimize the delay from input speech to output text.
Moreover, the size of the model cannot be too large as the inference speed and hardware resource usage are critical considerations for on-device use cases.
Due to these constraints, streaming models often exhibit lower recognition accuracy compared to non-streaming (a.k.a., full-context) models.

\begin{figure}[t]
    \centering
    \vspace{0.2cm}
    \resizebox{1.0\linewidth}{!}{
    \includegraphics{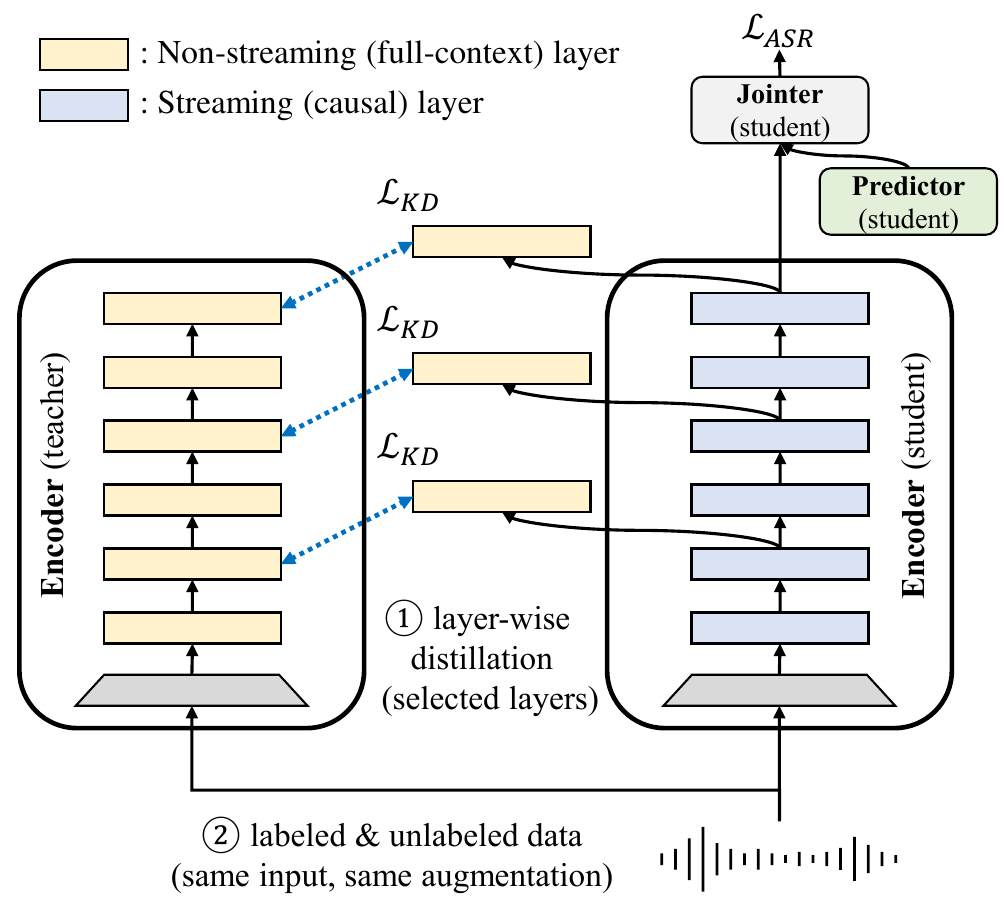}}
    \caption{Overview of the proposed knowledge distillation (KD) from non-streaming ASR teacher to streaming ASR student. Auxiliary non-streaming (yellow) layers are attached for KD.}
    \label{fig:overview}
    \vspace{-0.3cm}
\end{figure}

Knowledge distillation (KD)~\cite{hinton2015distilling,li2014learning} has proved its effectiveness in reducing the performance gap between streaming and non-streaming models.
Specifically, KD-based methods leverage the non-streaming model as a teacher to transfer the learned knowledge to the streaming student model~\cite{kim2017improved}.
The simple and frequently used approach is to minimize Kullback-Leibler divergence (KLD) between the output token emission probabilities (i.e., posterior distributions) of the teacher and student.
However, several studies have reported that direct frame-to-frame distillation could misguide the streaming model~\cite{yu2021dualmode,inaguma2021alignment,liang2022fastu2}.
This is because the audio-text frame alignments are different between the non-streaming and streaming models; the former generally emit tokens faster than the latter.
On the other hand, recent studies have introduced dual-mode~\cite{yu2021dualmode}, a type of self-KD training, which shares the same model for both streaming and non-streaming cases. 
While this approach improves streaming ASR performance, there exists a critical limitation that the student model size should be the same as the teacher~\cite{liu2023dualpruning}.

\begin{figure*}[t]
    \centering
    \resizebox{1.0\linewidth}{!}{
    \includegraphics{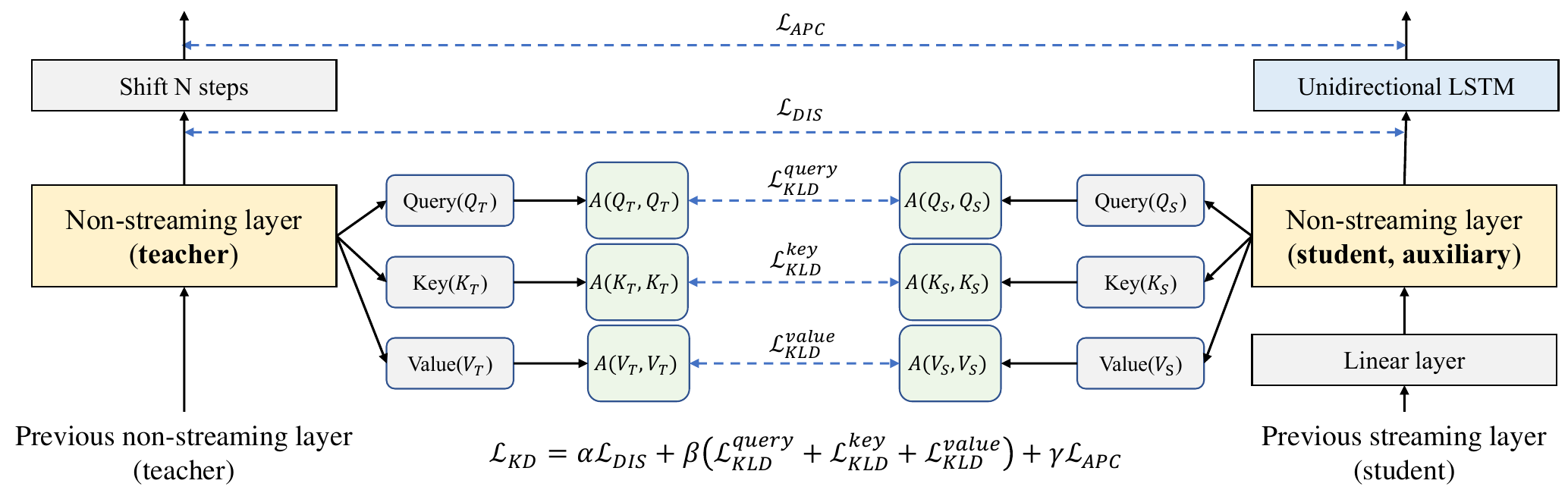}}
    \caption{Illustration of three distillation loss functions: feature distance error (DIS), Kullback–Leibler divergence (KLD) for query, key, value relationships, and auto-regressive predictive coding (APC) error. $\alpha$, $\beta$, and $\gamma$ are the weighting coefficients.}
    \label{fig:distillation}
    \vspace{-0.2cm}
\end{figure*}

In this work, we propose a novel KD method that applies layer-wise distillation on the encoder part of the model.
The proposed method offers several advantages over existing methods, including faster training, better robustness to frame misalignment, and the ability to utilize unlabeled data.
First, we insert auxiliary non-streaming layers into the streaming model (see Figure~\ref{fig:overview}).
These layers allow both the source and target of KD to observe the entire sequence, thereby minimizing discrepancies in terms of accessible context and making the training easier.
Second, we design a special KD loss function to enhance the feature extraction process (see Figure~\ref{fig:distillation}).
Our KD loss is composed of three losses targeting different objectives: 1) reducing the difference in the extracted feature of each frame, 2) matching the frame-to-frame relationship between all frames, and 3) predicting the future from the past context.
Especially, the third loss is inspired by autoregressive predictive coding (APC), which was originally suggested for self-supervised learning tasks.
We empirically discover that this \textit{future prediction} task is very helpful for streaming models.

Experimental results show that the proposed KD method considerably improves the streaming ASR performance.
Compared to the baseline, our method achieves 16\% relative WER reduction on LibriSpeech \textit{test-other} dataset.
By incorporating the large unlabeled data as an additional training resource, we could additionally reduce the WER by 3\%.

\section{Related Work}\label{sec:related}

\subsection{Distillation for Improving Streaming ASR}

KD has been widely used for improving the streaming ASR model by setting the non-streaming ASR model as a teacher.
The most popular approach is to distill the output token probabilities~\cite{li2014learning,inaguma2021alignment,kurata2020knowledge}.
To mitigate the alignment mismatch problem, previous studies suggested manual shifting~\cite{yang2022knowledge,weninger2022conformer}, multi-stage alignment matching~\cite{kurata2020knowledge}, or exploiting CTC alignments~\cite{inaguma2021alignment}.


Another line of research called dual-mode, where the non-streaming teacher and streaming student share the same parameter set, has been proposed~\cite{yu2021dualmode}.
Dual-mode can provide a considerably better performance, but this may be not the best fit for low-resource streaming ASR: 1) the teacher and the student should be the same (large) architecture, and 2) it has not been tested for chunk-wise streaming setting.
The problems of dual-mode have partially solved by joint training and student pruning together~\cite{liu2023dualpruning}, but this method exploits block sparsity that has limitations on most hardware platforms.


Recently, a layer-wise KD method has been introduced~\cite{kojima21transducer}.
However, this work is very distinct from ours; the main difference is the use of auxiliary non-streaming layers.
The method directly minimizes the frame-wise feature differences between the streaming and non-streaming layer, and this method requires the teacher and student to use the same feature dimension.
In addition, our KD loss can transfer diverse characteristics using three different loss objectives.


\subsection{Distillation for ASR Model Compression}

ASR models have adopted KD for model compression as KD's original purpose~\cite{hinton2015distilling}.
In particular, Transducer-based end-to-end ASR models~\cite{graves2012transducer} have been actively studied for compression~\cite{zhao22freplacing,rathod22multistage,park2023compression,swaminathan2021codert,panchapagesan2021efficient} because the framework is suitable for low-resource scenarios: on-device and real-time inference.
In addition, large self-supervised models have also tried KD to squeeze their model size~\cite{chang2022distilhubert}.


\subsection{Cascaded Non-streaming and Streaming Layers}

Cascade encoders are a special encoder design that stack several non-streaming layers on top of the streaming layers~\cite{li2021cascade,sainath2022improving}.
By doing so, the model can be jointly trained for non-streaming and streaming purposes, and the model can dynamically select which path to pass through~\cite{ding2022unified}.
Our work also attaches additional non-streaming layers, but they are only used for KD purpose and then removed after training is finished.


\section{Distillation from Non-streaming Teacher to Streaming Student}\label{sec:method}

\subsection{Preliminaries on ASR Encoder}\label{ssec:method_preliminaries}

Given the input speech frames $\mathbf{x}=\{x_1, x_2, ..., x_T\}$ and the target text tokens $\mathbf{y}=\{y_1, y_2, ...,  y_U\}$, the goal of ASR model is to maximize the likelihood $P(\mathbf{y}|\mathbf{x})$.
The input $\mathbf{x}$ first passes through an encoder, which is usually a stack of identical layers, to extract contextualized features for each frame.
Let the output feature of the $\ell{\text{-th}}$ layer $\mathbf{h}^{\ell} =\{h^{\ell}_1, h^{\ell}_2, ... , h^{\ell}_T\}$, and then the next layer's $t{\text{-th}}$ frame output can be represented as: $h^{\ell+1}_t = f^{\ell+1}(h^{\ell}_{t-LC}, ..., h^{\ell}_{t+RC})$ and $f^{\ell+1}(h^{\ell}_1, ..., h^{\ell}_T)$ for streaming and non-streaming models, respectively. 
LC and RC are the left/right-context size, where the right-context size is often set to zero in low-latency use cases.
For clarity, we will use $\mathbf{h}^{\ell}$ for the teacher and $\mathbf{s}^{\ell}$ for the student.
In this work, we employ Transducer~\cite{graves2012transducer} architecture; the encoder output is jointly combined with the predictor output and generates a token emission probability matrix of size $T\times U$.

\subsection{Layer-wise Distillation using Auxiliary Layer}\label{ssec:method_layerwise}

Instead of the output token probability distillation, we propose a layer-wise KD from the teacher encoder to the student encoder.
We conjecture that the naive frame-wise feature mapping from $h^{\ell}_t$ to $s^{\ell}_t$ would be unnecessarily difficult for the student model because of the very different context information teacher and student use.
The observation that non-streaming and streaming models produce different audio-text alignments can be strong evidence of our hypothesis.

In order to solve this context mismatch, we insert auxiliary non-streaming layers at selected layers of the student (see Figure~\ref{fig:overview}).
Formally, if an auxiliary layer is attached at $\ell{\text{-th}}$ layer, the output frame features of auxiliary layer $z^{\ell}_t = g^{\ell}(s^{\ell}_1, ..., s^{\ell}_T)$ can observe the entire sequence.
Then, we perform KD between the teacher layer $\mathbf{h}^{\ell}$ and the auxiliary layer $\mathbf{z}^{\ell}$.
By doing so, we can remove the fundamental context difference between the non-streaming source and streaming target, thus expecting the student to more focus on improving the quality of features.

The auxiliary layer consists of three parts (see Figure~\ref{fig:distillation}).
The feature first passes through a simple projection layer to expand the feature dimension the same as the teacher.
Then, it passes through a single Transformer layer and finally passes through a unidirectional LSTM layer.
Note that the auxiliary layers are jointly trained with the student encoder, while the teacher is fixed during training.

\subsection{Distillation Loss}\label{ssec:method_loss}

We propose a novel loss function designed for KD from the non-streaming ASR teacher to the streaming ASR student.

\subsubsection{Feature similarity loss}~\label{sssec:dis_loss}
First, we minimize the distance between the teacher/student features of the same frame to guide the feature extraction.
We adopt the distance function introduced in DistilHuBERT~\cite{chang2022distilhubert}.
\begin{equation}
    L_{\text{DIS}} = \sum_{t=1}^{T} \Big[ \frac{1}{D} || h_t - z_t ||_1 - \log \sigma \big( \cos (h_t, z_t)  \big)    \Big]
\end{equation}
where $D$ is the feature vector dimension, $\sigma$ is sigmoid activation function, and $\cos( \cdot,\cdot )$ is cosine similarity.

\subsubsection{Self-Attention loss}\label{sssec:kld_loss}
Second, we also distill the self-attention distribution to not only mimic the teacher frame-wise but also follow the inter-frame relationships.
Specifically, we transfer three relationships (Query-Query, Key-Key, and Value-Value) individually as suggested in MiniLM-v2~\cite{wang2021minilmv2}.
For simplicity,  we only present the Query-Query relationship below, but the equation is easily expandable for Key and Value cases.
\begin{equation}
    L_{\text{KLD}}^{\text{query}} = \frac{1}{A} \sum_{a=1}^{A} \sum_{t=1}^{T} \text{KLD} \big( R_{T(a,t)}^{\text{query}} || R_{S(a,t)}^{\text{query}}  \big)
\end{equation}
where $A$ is the number of attention heads.
$R_{T(a,t)}, R_{S(a,t)}$ indicate teacher/student attention probability distribution of the $t\text{-th}$ frame computed at $a\text{-th}$ attention head.
\begin{equation}
    R_{(a,t)}^{\text{query}} = \text{Softmax}_{k} \Big( q_{(a,t)} q_{(a,k)}^T  / \sqrt{d_A}  \Big)
\end{equation}
where $d_A=D/A$ is the attention head dimension and $q_{(a, t)}$ is a query vector of $t\text{-th}$ frame.
The self-attention loss is the sum of losses computed from Query, Key, and Value as:
\begin{equation}
    L_{\text{KLD}} = L_{\text{KLD}}^{\text{query}} + L_{\text{KLD}}^{\text{key}} + L_{\text{KLD}}^{\text{value}}
\end{equation}

\subsubsection{Future prediction loss}\label{sssec:apc_loss}
Finally, we employ the future prediction loss, inspired by the autoregressive predictive coding (APC)~\cite{chung2020apc,chung20vqapc}.
The objective of APC is to predict the future feature (after several frames) using previous history.
To do so, APC includes a unidirectional RNNs and train those layers to minimize the difference.
Although APC was originally proposed for self-supervised speech pre-training, we discover that APC can be especially helpful for streaming ASR model distillation.
Let the output features of the unidirectional LSTM layer $r_t = \text{LSTM}(z_1, z_2, ... z_t)$ (see Figure~\ref{fig:distillation}), then our APC loss is computed as:
\begin{equation}
    L_{\text{APC}} = \sum_{t=1}^{T} \Big[ \frac{1}{D} || h_{t+N} - r_t ||_1 - \log \sigma \big(  \cos (h_{t+N}, r_t)  \big)  \Big]
\end{equation}
where $N$ indicates how far the target frame is located.
Note that the equation is almost the same to the feature similarity loss, but the target is shifted by $N$ frames.
APC loss encourages the model to predict future, trying to follow the teacher's behavior as if the student can exploit the future context.

\begin{figure}[t]
    \centering
    \resizebox{0.82\linewidth}{!}{
    \includegraphics{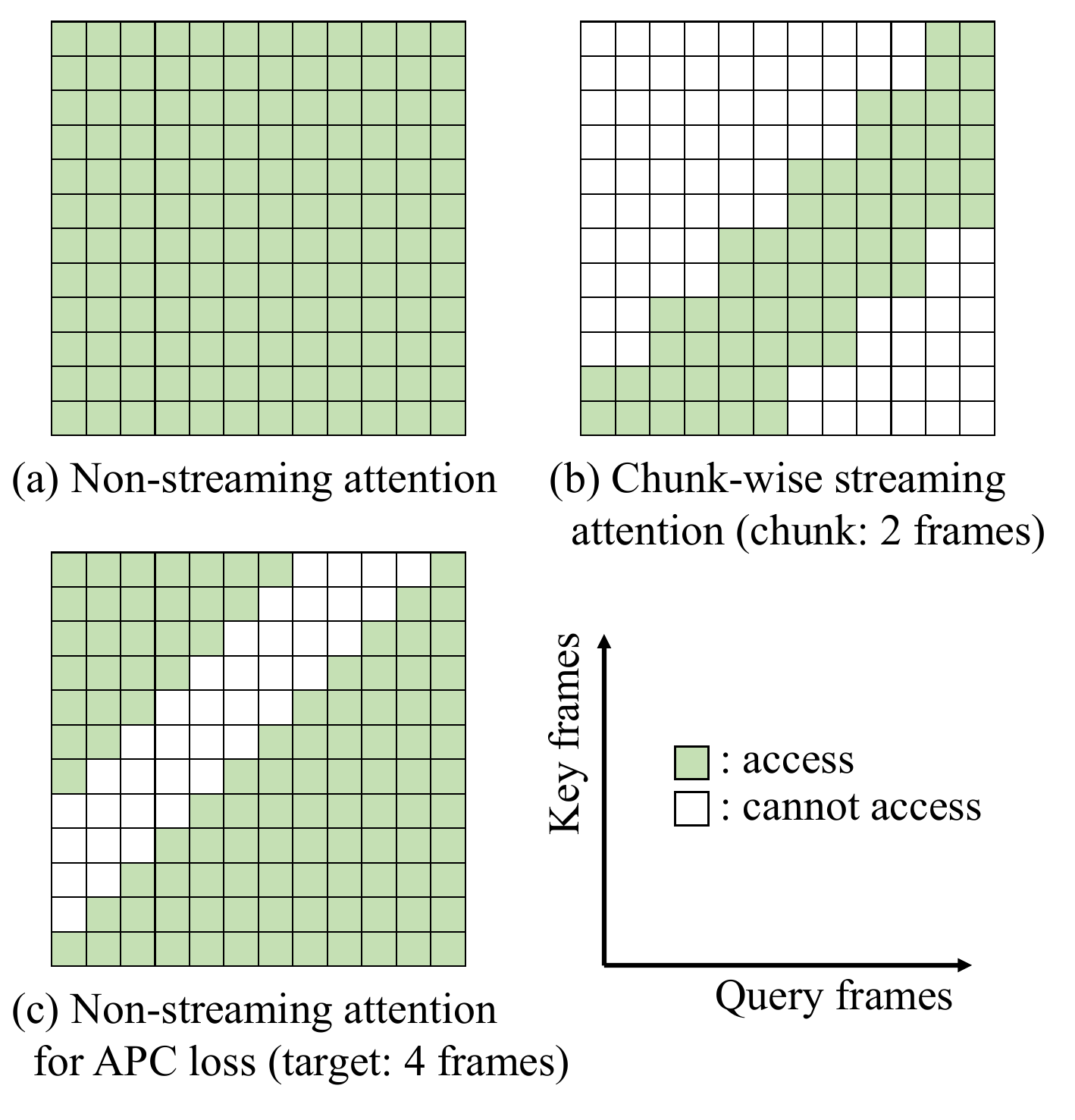}}
    \caption{Illustration of self-attention masks in (a) non-streaming layer in teacher, (b) streaming layer in student, and (c) non-streaming layer in auxiliary branches. Note that the number of frames in the figure is chosen for clear understanding.}
    \vspace{-0.2cm}
    \label{fig:attention_mask}
\end{figure}

To maximize the efficiency of APC loss, we modify the Transformer attention mask applied for auxiliary non-streaming layers (see Figure~\ref{fig:attention_mask}(c)).
In particular, we mask future $N$ frames $[t+1, t+N]$ so that each frame cannot directly observe frames between the current one and the target of APC loss.
By doing so, we can prevent the model from cheating (a.k.a., `information leakage') by directly taking the target from the input sequence.

\subsubsection{Total loss}\label{sssec:total_loss}

Overall, the final loss function is combined as follows:
\begin{align}
    &L_{\text{total}} = \frac{1}{N_s} \sum_{i\in \{N_s\}} L_{\text{ASR}}(\textbf{x}_i) \\
    &+ \frac{1}{N_s + N_u} \sum_{i\in \{N_s,N_u\}} \Big( \alpha L_{\text{DIS}} + \beta  L_{\text{KLD}} + \gamma L_{\text{APC}} \Big )(\textbf{x}_i) \nonumber 
\end{align}
where $N_s$, $N_u$ are the number of labeled and unlabeled samples, respectively.
We set $\alpha=0.01$, $\beta=0.0005$, and $\gamma=0.005$ for the experiments.

\section{Experimental Results}\label{sec:experiment}

\subsection{Setup}\label{ssec:experiment_setup}

\vspace{0.1cm}
\noindent\textbf{Dataset.} 
We use LibriSpeech-960h~\cite{panayotov2015librispeech} dataset for supervised ASR training.
Additionally, we exploit 6K hours of unlabeled data from the same domain (i.e., audio book reading)~\cite{kahn2020librilight} to better transfer knowledge from the teacher encoder.

\vspace{0.2cm}
\noindent\textbf{Model Architecture.} 
In this work, we employ the Conformer-Transducer~\cite{gulati20conformer} ASR models.
The teacher encoder is a stack of 16 Conformer layers with 512-dim. feature and 8 attention heads.
The student encoder is a stack of 16 Conformer layers with 256-dim. feature and 4 attention heads.
For the streaming student, the convolution layer is set to be causal and the attention mask is restricted to only access chunks within the left context limit.
For the auxiliary layers, we use a single Transformer layer (self-attention + feed-forward).
We apply KD from the $4, 8, 12, 16\text{-th}$ layers of the teacher to $4, 8, 12, 16\text{-th}$ layers of the student, respectively.

\vspace{0.2cm}
\noindent\textbf{Training Details.} 
We basically follow the training setup of the original Conformer paper~\cite{gulati20conformer} for student training, using a batch size of 1024 and peak learning rate of 1.25e${^{-3}}$.
The streaming student uses a chunk-wise training and inference~\cite{chen2021developing}, where the chunk size is 160ms and the left/right context size is 640ms and 0ms, respectively (see Figure~\ref{fig:attention_mask}(b)).
When unlabeled data is used, we mix the labeled and unlabeled data to a 1:1 ratio to construct an epoch.

\subsection{ASR Performance}\label{ssec:experiment_performance}

\setlength{\tabcolsep}{8pt}
\begin{table}[t]
    \centering
    \caption{
    Word error rate (\%) for three streaming student models (S1-S3) and two small models (NS and S0) trained from scratch. The teacher model is a non-streaming model.}
    \vspace{-0.1cm}
    \resizebox{1.0\linewidth}{!}{
    \begin{tabular}{l|rrrr}
        \toprule
        \multirow{2}{*}{Model}  & \multicolumn{2}{c}{\textit{dev-}} & \multicolumn{2}{c}{\textit{test-}} \\
                                &   \textit{clean} & \textit{other} & \textit{clean} & \textit{other} \\
         \midrule
         Teacher (106M)                 & 2.2  & 5.3  & 2.5  & 5.4 \\
         \midrule
         From scratch (30M) & & & & \\
         - NS (non-streaming)               & 2.4   & 5.7   & 2.7   & 5.9 \\
         - S0 (streaming)                   & 4.9   & 12.2  & 4.7   & 12.8 \\
         \midrule
         Distillation (30M) & & & & \\
         - S1 (prob. KLD)                    & 4.8     & 12.0     & 4.4     & 13.3 \\
         - S2 (KD w/o aux. layer)            & 4.0     & 11.1     & 4.4     & 12.3 \\
         - \textbf{S3 (KD w/ aux. layer)}    & \textbf{3.8}     & \textbf{10.1}     & \textbf{4.1}     & \textbf{10.7} \\
        \bottomrule
    \end{tabular}}
    \label{tab:wer}
    \vspace{0.1cm}
\end{table}

\setlength{\tabcolsep}{7pt}
\begin{table}[t]
    \centering
    \caption{Ablation study on the knowledge distillation losses. $N$ is the relative target distance in APC loss.}
    \vspace{-0.1cm}
    \resizebox{1.0\linewidth}{!}{
    \begin{tabular}{ccc|rrrr}
        \toprule
        \multicolumn{3}{c|}{Loss ablation}    & \multicolumn{2}{c}{\textit{dev-}} & \multicolumn{2}{c}{\textit{test-}} \\
        DIS & KLD & APC             &   \textit{clean} & \textit{other} & \textit{clean} & \textit{other} \\
        \midrule
        \checkmark   &  &                                   & 4.0     & 10.8     & 4.3     & 11.6 \\
        \checkmark   & \checkmark &                         & 4.0     & 10.7     & 4.2     & 11.2 \\
        \checkmark   & \checkmark & \checkmark ($N$=2)      & 3.9     & 10.4     & 4.1     & 10.9 \\
        \checkmark   & \checkmark & \checkmark ($N$=4)      & \textbf{3.8}     & \textbf{10.1}     & \textbf{4.1}     & \textbf{10.7} \\
        \checkmark   & \checkmark & \checkmark ($N$=8)      & 4.0     & 10.5     & 4.2     & 11.2 \\
        \bottomrule
    \end{tabular}}
    \label{tab:ablation}
    \vspace{-0.2cm}
\end{table}

Table~\ref{tab:wer} compares WER of different training methods on LibriSpeech evaluation datasets.
Note that the external language model is not used during decoding.
S1 is the model trained with token probability distillation, and S2 is the case that directly applies $L_{\text{DIS}}$ for layer-wise KD without the proposed auxiliary layer.
We can observe that the proposed training method (S3) achieves the lowest WER.
Interestingly, the conventional method (S1) only achieves a marginal improvement compared to the model without KD (from scratch, S0).
We would like to mention that we have tested various hyper-parameter configurations for S1 to stabilize the training and improve performance.
We are cautiously assuming that the method would not be the best fit for our target setup, small-size model and chunk-wise streaming.

\subsection{Ablation Study}\label{ssec:experiment_ablation}

Table~\ref{tab:ablation} shows that all three KD losses contribute to the performance.
For APC, we test three $N$ configurations (see Section~\ref{sssec:apc_loss}) and empirically find that the best results are produced with $N$=4, which corresponds to 160ms of speech.
Furthermore, by comparing S2 (Table~\ref{tab:wer}) and the first row (Table~\ref{tab:ablation}) that use the same loss, we can verify that the auxiliary non-streaming layer indeed plays an important role.

\subsection{Additional Unsupervised Training}\label{ssec:experiment_more_data}

\setlength{\tabcolsep}{9pt}
\begin{table}[t]
    \centering
    \caption{Effect of the size of unlabeled data for the distillation.}
    \vspace{-0.1cm}
    \resizebox{1.0\linewidth}{!}{
    \begin{tabular}{l|rrrr}
        \toprule
        \multicolumn{1}{c|}{\#Unlabeled}   & \multicolumn{2}{c}{\textit{dev-}} & \multicolumn{2}{c}{\textit{test-}} \\
        \multicolumn{1}{c|}{data (hrs)}       &   \textit{clean} & \textit{other} & \textit{clean} & \textit{other} \\
        \midrule
        0 (none)            & 3.8     & 10.1     & 4.1     & 10.7 \\
        +600 (small)         & 3.8     & 9.8     & 4.2     & 10.4 \\
        +6,000 (medium)      & 3.8     & 9.7     & 4.4     & 10.3 \\
        \bottomrule
    \end{tabular}}
    \label{tab:dataset}
    \vspace{-0.3cm}
\end{table}

Table~\ref{tab:dataset} presents performance improvements when using 600 (small) and 6,000 (medium) hours of unlabeled data in LibriLight dataset~\cite{kahn2020librilight}.
When using 6,000 hours of unlabeled data, we observe a clear WER reduction on the challenging LibriSpeech \textit{dev-other} and \textit{test-other} datasets.
Nevertheless, we observe a slight WER increase in \textit{dev-clean} and \textit{test-clean}; we believe this is due to the imbalance of the number of labeled and unlabeled samples.
Because such large-scale data training takes much longer time, the models in Table~\ref{tab:dataset} started from the best model (S3) which is trained only with supervised data.
\section{Discussion}\label{sec:discussion}

\subsection{Teacher-to-Student Layer Connection}\label{ssec:discussion_layer_connection}

Although we connected the layers of the same position from both teacher and student, we expect more optimization can be possible if we could find the optimal connection between layers.
One approach is to identify the layers of similar behaviors and connect them.
Indeed, recent studies show that the information encoded in each layer can be examined, for both ASR models~\cite{shim2022understanding,shim22similarity} and self-supervised speech models~\cite{yang20iunderstanding,pasad2021layer}.
On the other hand, it would be an interesting research direction to reduce the number of layers in the student model for additional compression~\cite{chang2022distilhubert}.


\subsection{Relationship to Pseudo Label Generation}\label{ssec:discussion_pseudo_label}

To utilize unlabeled data, several studies have suggested generating pseudo labels for such data using the out-of-the-shelf large ASR model~\cite{hwang2022pseudo,yang2022knowledge}.
Previous work tested this idea for improving streaming ASR performance~\cite{doutre2021improving} and achieved considerable WER reduction, but only tested a large streaming model with more than 120M parameters.
We believe this approach can be combined with ours as they are orthogonal to each other.

\section{Conclusion}\label{sec:conclusion}

In this paper, we proposed a novel KD method for improving the small streaming student using the large non-streaming teacher.
The key idea is to insert non-streaming auxiliary layers on the student, which aims at matching the context of the target features for knowledge distillation.
We also designed a special layer-wise KD loss function, which includes future prediction task using APC loss.
Experiments showed that our method improves the streaming ASR performance by a clear margin.



\newpage
\bibliographystyle{IEEEtran}
\bibliography{reference}

\end{document}